\documentclass{llncs}
\usepackage{llncsdoc,graphicx}
\usepackage{amsmath}
\usepackage{multicol}
\usepackage{longtable}

\begin{document}

\title{A Tool for Computing and Estimating the Volume of the Solution Space of SMT(LA) Constraints}

\author{
Cunjing Ge,
Feifei Ma
 and
Jian Zhang
}
\institute{
State Key Laboratory of Computer Science\\
Institute of Software, Chinese Academy of Sciences\\
Email: ~\{gecj,maff,zj\}@ios.ac.cn
}

\maketitle

\begin{abstract}
There are already quite a few tools for solving the Satisfiability Modulo Theories (SMT) problems.
In this paper, we present \texttt{VolCE}, a tool for counting the solutions of SMT constraints, or
in other words, for computing the volume of the solution space. Its input is essentially
a set of Boolean combinations of linear constraints, where the numeric
variables are either all integers or all reals, and each variable is bounded.
The tool extends SMT solving with
integer solution counting and
volume computation/estimation for convex polytopes.
Effective heuristics are adopted, which enable the tool to deal with
high-dimensional problem instances efficiently and accurately.
\end{abstract}

\section{Introduction}

In recent years, there have been a lot of works on solving
the Satisfiability Modulo Theories (SMT) problem.
Quite efficient SMT solvers have been developed, such as CVC3, MathSAT, Yices and Z3.
In \cite{MLZ09}, we studied the counting version of SMT solving, and
presented some techniques for computing the size of the solution space
efficiently. This problem can be regarded as an extension to SMT solving,
and also an extension to model counting in the propositional logic.
It has recently gained much attention in the software engineering community
\cite{LZ11,GDV12}.

The prototype tool presented in \cite{MLZ09} computes the exact volume of solution space.
However, exact volume computation in general is an extremely difficult problem.
It has been proved to be \#P-hard, even for explicitly described polytopes.
On the other hand, it suffices to have an approximate value of the volume in many cases.
Recently we implemented a tool to estimate the volume of polytopes;
and integrated it into the framework of \cite{MLZ09}.

This paper presents the new tool \texttt{VolCE}\footnote{It is available at
{\tt http://lcs.ios.ac.cn/}\~{\tt zj/volce10x64.tar.gz}} for the counting version of
SMT(LA). (Here LA stands for linear arithmetic.)
The input of the tool is a set of Boolean combinations of linear constraints, where
each numeric variable is bounded.
Independent Boolean variables may also appear in the constraints.
The output of the tool is the ``volume'' of the solution space, or the number of solutions
in case that the domain consists of integer points.

The rest of this paper is organized as follows.
Section~\ref{sect:Arch} presents the architecture of \texttt{VolCE}.
Section~\ref{sect:vol} presents the algorithm and implementation of our volume estimation sub-procedure.
Section~\ref{sect:usage} briefly describes how to use the tool. 
Section~\ref{sect:exps} discusses our experiments, and
Section~\ref{sect:reltwks} describes some related works.
We conclude in Section~\ref{sect:conclude}.

\section{Architecture of \texttt{VolCE}}\label{sect:Arch}

\begin{figure}
\centering
\includegraphics[scale=0.6]{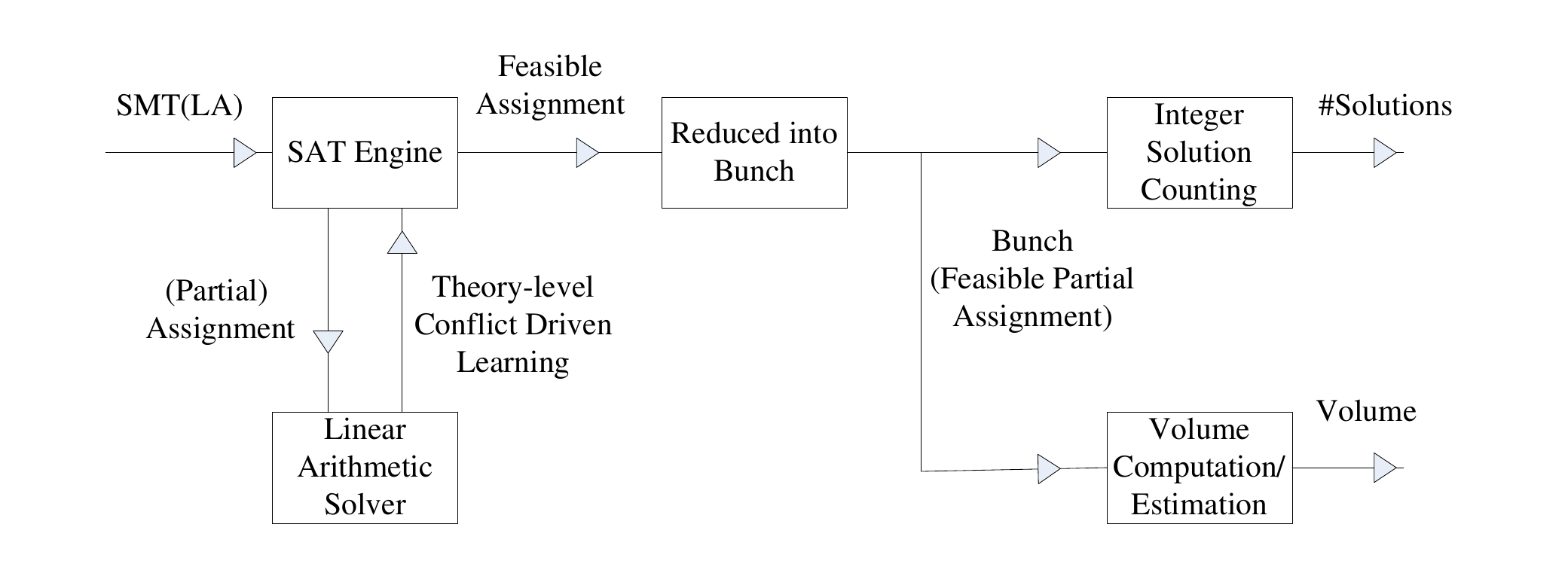}
\caption{The Architecture of \texttt{VolCE}}
\label{volcepp}
\end{figure}

The architecture of \texttt{VolCE} is illustrated in Figure~\ref{volcepp}. Recall that in \cite{MLZ09},
a consistent conjunction of linear constraints that satisfies the boolean skeleton of the SMT(LA) formula
is called a feasible assignment. The sum of volumes of all feasible assignments is the volume of
the whole formula. In \texttt{VolCE}, the SAT solving engine ({\tt MiniSat}\footnote{
The MiniSat Page. {\tt http://minisat.se/}})
and the linear arithmetic solver ({\tt lp\_solve}\footnote{Available at {\tt http://lpsolve.sourceforge.net/}})
work together to find feasible assignments.
Each time a feasible assignment is obtained, \texttt{VolCE} tries to reduce it to
a partial assignment that still propositionally satisfies the formula.
The resulted feasible partial assignment may cover a bunch of feasible assignments,
hence is called a ``bunch".
Then a solution counting or volume computation/estimation sub-procedure is called
for the polytope corresponding to each bunch rather than each feasible assignment,
so that the number of calls is reduced.
For more details of the main algorithm, see~\cite{MLZ09}.

In addition to {\tt MiniSat} and {\tt lp\_solve},
\texttt{VolCE} calls {\tt Vinci}~\cite{vinci98} and {\tt LattE}~\cite{latte04} to help compute
the size of the solution space.
\texttt{Vinci} is a tool for computing the volume of a convex body.
{\tt LattE} is a tool for counting lattice points inside convex polytopes and solutions of integer programs.
Moreover, we implemented a tool for estimating the volume of
convex polytopes, called \texttt{PolyVest}~\cite{polyvest}.
It will be elaborated in the subsequent section.

\section{Volume Estimation}\label{sect:vol}
The performance of volume computation packages for convex polytopes is the bottleneck of the prototype tool in~\cite{MLZ09}.
Recently we augmented it with an efficient volume estimation sub-procedure for convex polytopes.

\subsection{Volume Estimation for Convex Polytopes}
A straightforward way to estimate the volume of a convex body is the Monte-Carlo method.
However, it suffers from the curse of dimensionality\footnote{http://en.wikipedia.org/wiki/Curse\_of\_dimensionality},
which means the possibility of sampling inside a certain space in the target object decreases very quickly while the dimension increases.
As a result, the sample size has to grow exponentially to achieve a reasonable estimation.
To avoid the curse of dimensionality, Dyer et.al. proposed a polynomial time randomized approximation algorithm
(Multiphase Monte-Carlo Algorithm)~\cite{Kannan:1989}.
The theoretical complexity of the original algorithm is $O^*(n^{23})$\footnote
{``soft-O'' notation $O^*$ indicates that we suppress factors of $\log n$ as well as factors depending on other parameters like the error bound},
and is recently reduced to $O^*(n^4)$~\cite{Lovasz:2006}.

Based on the Multiphase Monte-Carlo method,
we implemented our own tool \texttt{PolyVest} (Polytope Volume Estimation) to estimate the volume of convex polytopes~\cite{polyvest}.
One improvement over the original Multiphase Monte-Carlo method is that we developed a new technique to reutilize sample points,
so that the number of sample points can be significantly reduced.

In the sequel, we briefly describe the algorithm implemented in \texttt{PolyVest}.
For more details, one can refer to~\cite{polyvest}.
We assume that $P$ is a full-dimensional and nonempty convex polytope.
We use $vol(K)$ to represent the volume of a convex body $K$, and $B(x, R)$ to represent the ball with radius $R$ and center $x$.

\begin{figure}[htbp]
\centering
\begin{minipage}[t]{0.45\textwidth}
\centering
\includegraphics[width=4cm]{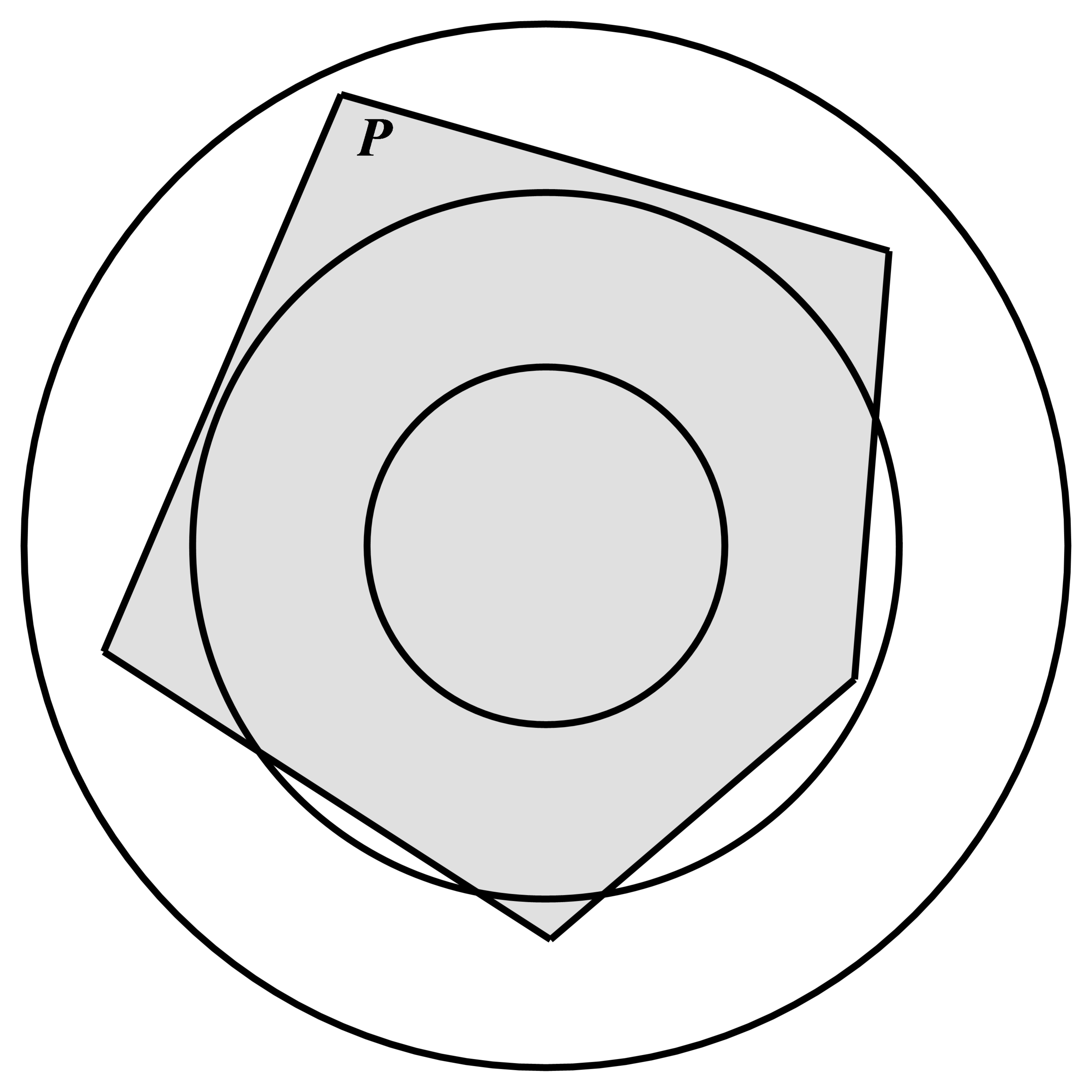}
\caption{Multiphase Monte-Carlo}\label{mmc}
\end{minipage}
\begin{minipage}[t]{0.45\textwidth}
\centering
\includegraphics[width=4cm]{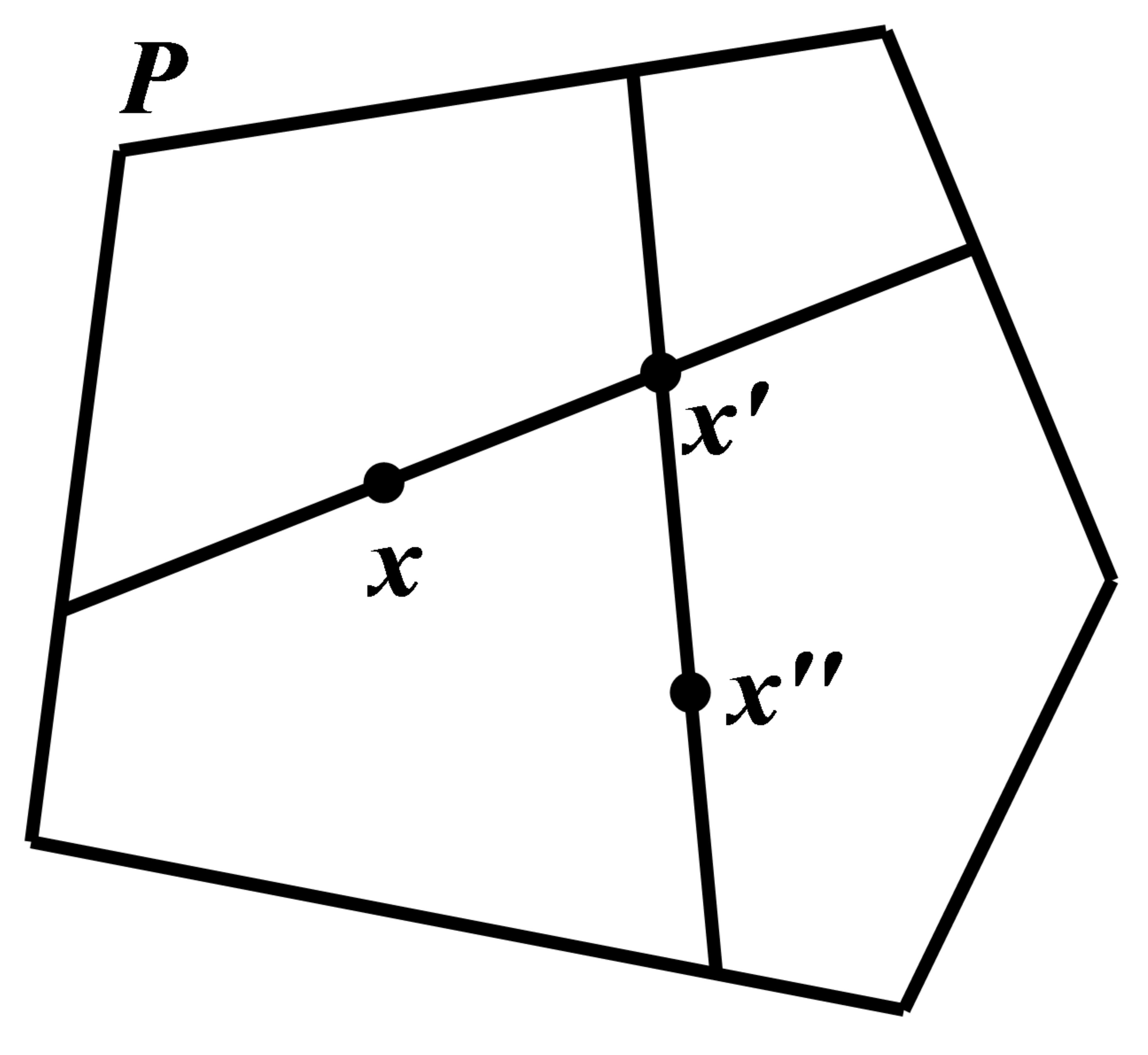}
\caption{Hit-and-run}\label{hnr}
\end{minipage}
\end{figure}

The basic procedure of \texttt{PolyVest} consists of the following three steps: rounding, subdivision and sampling.

{\bf Rounding}~
First we find an affine transformation $T$ on polytope $P$
so that $T(P)$ contains the unit ball $B(0, 1)$, and is contained in the ball $B(0, r)$.
This can be achieved by applying the Shallow-$\beta$-Cut Ellipsoid Method~\cite{Book:1993}.
We set $r=2n$ in our implementation.
Rounding is an essential step.
For example, it is difficult to subdivide a very ``thin'' polytope and sample on it without rounding.
For simplicity, we still use $P$ to denote the new polytope $T(P)$ after rounding.

{\bf Subdivision}~
Then we divide $P$ into a sequence of convex bodies.
The general idea of the subdivision step is illustrated in Figure~\ref{mmc}.
We place $l$ concentric balls $\{B_i\}$ between $B(0, 1)$ and $B(0, r)$.
Let $K_i$ denote the convex body $B_i \cap P$, then
\begin{displaymath}
vol(P) = vol(K_0)\prod_{i=0}^{l-1} \frac{vol(K_{i+1})}{vol(K_i)}.
\end{displaymath}
Let $\alpha_i$ denote the ratio $vol(K_{i+1})/vol(K_i)$, then
\begin{displaymath}
vol(P) = vol(K_0)\prod_{i=0}^{l-1} \alpha_i.
\end{displaymath}
Hence the volume of the polytope $P$ is transformed to the products of ratios and the volume of $K_0$.
Note that $K_0 = B(0, 1)$ whose volume can be easily computed.
So we only have to estimate the value of $\alpha_i$.

Of course, one would like to choose the number of concentric balls, $l$, to be small.
However, one needs about $O(l\alpha_i)$ random points to get a sufficiently good approximation for $\alpha_i$.
It follows that the $\alpha_i$ must not be too large.
In \texttt{PolyVest}, we set $l = \lceil n\log_2r \rceil$ and $B_i = B(0, 2^{i/n})$ to construct the convex bodies $\{K_i\}$.
And it can be proved that $1\le\alpha_i\le 2$ with this construction.

{\bf Sampling}~
Finally, we generate $S$ points in $K_{i+1}$ and count the number of points $c_i$ in $K_i$.
Thus $\alpha_i$ can be approximated with $S / c_i$.
Generating independent uniformly distributed random points in $\{K_i\}$ is not as simple as in cubes or ellipsoids.
So we use a hit-and-run method for sampling.
Hit-and-run method is a random walk which can generate points with almost uniform distribution in polynomial time~\cite{Smith:1987}.
Figure~\ref{hnr} illustrates the hit-and-run method:
It starts from a point $x$, then randomly selects a line $L$ through $x$ and choose the next point $x'$ uniformly on the segment in $P$ of line $L$.
\texttt{PolyVest} adopts the coordinate directions hit-and-run method,
in which the random direction of line $L$ is chosen with equal probability from the coordinate direction vectors and their negations.

\emph{Reutilization of Sample Points}~
In the original Multiphase Monte Carlo method,
the ratios $\alpha_i$ are estimated in natural order,
from the first ratio $\alpha_0$ to the last one $\alpha_{l-1}$.
And the method starts sampling from the origin.
However, our implementation works in the opposite way.
It generates sample points from the outermost convex body $K_l$ to the innermost convex body $K_0$,
and the ratios are estimated accordingly in reverse order.

The advantage of approximation in reverse order is that it is possible to fully exploit the sample points generated
in the previous phases.
Since $K_0 \subseteq K_1 \subseteq \dots \subseteq K_l$,
the sample points in $K_i$ still fall in $K_j$ ($i \le j$).
On the other hand, the sample points generated by the hit-and-run method are almost independent as sample size $S$ is large enough.
Therefore, for any $i$ that $i \le j$,
the points generated for approximating $\alpha_j$ that hit $K_{i+1}$ can serve as sample points to approximate $\alpha_i$ as well.
It can be easily proved that we only need to generate less than half sample points with this technique since $\alpha_i \le 2$.
In practice, this technique can save over 70\% time consumption under most circumstances.

\subsection{Volume Estimation for SMT(LRA) Formula}

Now we describe how to estimate the ``volume'' of the solution space of SMT(LRA) formulas.
(Here LRA stands for linear \emph{real} arithmetic.)
The basic procedure is quite similar to that of volume computation as described in~\cite{MLZ09}.
As in~\cite{MLZ09}, a consistent conjunction of linear constraints that satisfies
the boolean skeleton of the SMT(LRA) formula is called a feasible (partial) assignment.
Each time \texttt{VolCE} obtains a feasible (partial) assignment,
it calls \texttt{PolyVest} to estimate the volume of the polytope corresponding to this assignment.
The sum of estimated volumes of all feasible (partial) assignments is approximately the volume for the whole formula.
Note that the ``volume computation in bunches" strategy in~\cite{MLZ09} can also be applied in volume estimation.

In the Multiphase Monte-Carlo method, the number of sample points at each phase is a key parameter.
As the sample size increases, the accuracy of estimation improves, and the estimation process also takes more time.
It is important to balance the accuracy and run time, especially for \texttt{VolCE} since the estimation subroutine \texttt{PolyVest} is usually called many times.

\texttt{VolCE} employs a two-round strategy that can dynamically determine a proper sample size for each feasible (partial) assignment.
At the first round of estimation, each feasible assignment is sampled with a fixed small number of random points to get a quick and rough estimation.
Since the volumes of feasible assignments may vary a lot,
intuitively a feasible assignment with relatively larger volume should be estimated with higher accuracy.
Hence at the second round, the sample size for each assignment is determined according to its estimated volume from the first round.
More specifically, we use the following rule to decide the sample sizes in the second round:
\begin{itemize}
\item Suppose the sample size in the first round is $S_{min}$, and the largest sample size in the second round is set to $S_{max}$.
Let $V_{max}$ denote the largest estimated volume in the first round,
and $V_i$ denote the volume of the $i$th feasible assignment estimated in the first round.
Then the sample size $S_i$ for the $i$th feasible assignment in the second round is:
$$S_i = \frac{2\times S_{max}\times V_i}{V_{max}}$$
If $S_i\le S_{min}$, the $i$th feasible assignment is neglected at the second round,
and we use the result from the first round as its estimated volume.
If $S_i > S_{max}$, then set $S_i$ to $S_{max}$.
\end{itemize}
Through statistical results of substantial experiments,
we find that setting $S_{min}$ to $40l$ and $S_{max}$ to $1600l$ ($l = \lceil n\log_2r \rceil$) is very effective.
It only needs to generate $S_{min}/S_{max}=1/40$ points in extreme cases with this strategy.
In practice, it usually saves more than $95\%$ points for random instances.


\section{Using the Tool}\label{sect:usage}


The input of \texttt{VolCE} is an SMT formula where the theory $T$ is
restricted to the linear arithmetic theory.
It can be regarded as a Boolean combination of linear inequalities.
There are two input formats.

For the first format, the input formula is a Boolean
formula $\phi(b_1, \dots, b_n)$ in conjunctive normal form (CNF).
And each Boolean variable $b_i$ can stand for a linear arithmetic constraint (LAC).
The whole input file is an extension of the DIMACS
format for SAT solving. 
An alternative input format is SMT-Lib style. Currently,
\texttt{VolCE} supports the main features of the ``SMT-LIBv2'' syntax.


{\tt VolCE} has several command-line options:
\begin{itemize}
\item
\texttt{-V} asks the tool to call \texttt{Vinci} to compute the volume.
\item
\texttt{-P} asks the tool to call \texttt{PolyVest} to approximate the volume.
\item
\texttt{-L} asks the tool to call \texttt{LattE} to count integer solutions.
\item
\texttt{-w=NUMBER} specifies the word length of variables in bit-wise representations.
\item
\texttt{-maxc=NUMBER}
sets the maximum sampling coefficient of \texttt{PolyVest}, which is an upper bound.
\item
\texttt{-minc=NUMBER}
sets the minimum sampling coefficient of \texttt{PolyVest}.
\end{itemize}

For more details about using the tool, see the manual.
Here we just give an example to show its application to program analysis.

\subsection*{Example from Program Analysis}\label{sect:examples}
In~\cite{MLZ09}, we describe a program called {\it getop()}
and analyze the execution frequency of its paths.
For {\it Path1}, its path condition\footnote{The path condition
is a set of constraints such that any input data satisfying
these constraints will make the program execute along that path.}
is:
\begin{verbatim}
    (NOT ((c = 32) OR (c = 9) OR (c = 10))) AND
    ((c != 46) AND ((c < 48) OR (c > 57)))
\end{verbatim}
Here {\tt c} is a variable of type {\tt char}; it can be
regarded as an integer variable within the domain [-128..127].
For this path condition, we can compute the number of solutions
using \texttt{VolCE}.
With option {\tt -L} (i.e., using {\tt LattE}), the tool
tells us that the constraint has $242$ solutions.
(We do not need to use the option {\tt -w=8},
because the default word length is $8$.)
Given that the size of the whole search space is $256$, we conclude
that the frequency of executing {\it Path1} is about $0.945$ ($242/256$).
This means, if the input string has only one character, most probably,
the program will follow this path.

Another path, {\it Path2}, has
a more complicated path condition (omitted here, due to the lack of space).
It involves 3 input variables {\tt c0}, {\tt c1} and {\tt c2}.
Given the second path condition,
our tool tells us that the number of solutions is 8085.
So the path execution frequency is  $8085 / (256*256*256)$
which is roughly 0.00048.

\section{Experimental Results}\label{sect:exps}

In this section, we report some experimental results about our tool.
The experiments are performed on a workstation with 3.40GHz Intel Core i7-2600 CPU and 8GB memory.
In all experiments, the parameter $S_{max}$ is set to 1600, and $S_{min}$ set to 40.
The domain of each numeric variable is set to [-128..127] by default.

\subsubsection*{Benchmarks}
The following instances have been tested.
\begin{itemize}
\item
Instances generated from static program analysis. We analyzed the following programs:
\begin{itemize}
\item{abs}: a function which calculates absolute value;
\item{findmiddle}: a function which finds the middle number among 3 numbers;
\item{Space\_manage}: a program related to space technology;
\item{tritype}: a program which determines the type of a triangle;
\item{calDate}: a function which converts the special date into a Julian date;
\end{itemize}
\item
Instances from SMT-Lib, including: (1) the
QF\_LRA benchmarks {\tt Arthan, atan}; and (2) the QF\_LIA benchmarks
{\tt bignum, simplebitadder, fischer, pigeon-hole, prime\_cone}.
\item
Random instances {\tt ran}\_$i$\_$c$\_$d$: which have $d$ numeric variables, $i$ inequalities and $c$ clauses.
They are generated by randomly choosing coefficients of LACs and literals of clauses.
The length of each clause is between 3 to 5.
\end{itemize}

In the following tables, ``---'' means that the instance takes more than one hour to solve (or the tool runs out of memory).

\begin{table}[!htbp]
\footnotesize \centering
\setlength{\abovecaptionskip}{0pt}
\setlength{\belowcaptionskip}{5pt}
\caption{Comparison between Volume Estimation and Computation}\label{table:comp}
\newcommand{\tabincell}[2]{\begin{tabular}{@{}#1@{}}#2\end{tabular}}
\begin{tabular}{|c|c|c|c|c|c|c|c|c|}
\hline
\multicolumn{5}{|c|}{} & 	\multicolumn{2}{|c|}{PolyVest} & \multicolumn{2}{|c|}{Vinci} \\
\hline
Instance		& Dims	& Ineqs	& Clauses	& Bunches	& Result	& Time(s)	& Result	& Time(s)\\
\hline
abs\_0			& 1	& 1	& 1		& 1		& 127		& 0.000		& 127		& 0.000	\\
findmiddle\_2		& 3	& 6	& 20		& 2		& 5369040	& 0.016		& 5527125	& 0.000	\\
findmiddle\_3		& 3	& 6	& 20		& 7		& 5696270	& 0.020		& 5527125	& 0.004	\\
Space\_manage\_1	& 17	& 3	& 7		& 1		& 2.35e+39	& 2.6		& ---		& ---	\\
Arthan1A-chunk-0015	& 4	& 6	& 16		& 1		& 2.64e-3	& 0.016		& 2.57e-3	& 0.002 \\
 \tabincell{c}
{atan-problem-2-\\
weak-chunk-0200}	& 3	& 6	& 19		& 1		& 2.71		& 0.008		& 2.67		& 0.014 \\
ran\_15\_45\_7		& 7	& 15	& 45		& 113		& 1.84e+15	& 1.4		& 1.84e+15	& 12	\\
ran\_20\_60\_7		& 7	& 20	& 60		& 254		& 6.68e+14	& 2.75		& 6.74e+14	& 84	\\
ran\_30\_90\_7		& 7	& 30	& 90		& 401		& 4.62e+13	& 5.9		& 4.58e+13	& 802	\\
ran\_15\_45\_8		& 8	& 15	& 45		& 214		& 3.58e+17	& 2.9		& 3.50e+17	& 72	\\
ran\_20\_60\_8		& 8	& 20	& 60		& 480		& 1.07e+17	& 6.3		& 1.09e+17	& 259	\\
ran\_30\_90\_8		& 8	& 30	& 90		& 1135		& 6.73e+16	& 20.7		& ---		& --- 	\\
ran\_20\_60\_9		& 9	& 20	& 60 		& 425		& 1.20e+19	& 8.3		& ---		& ---	\\
ran\_20\_50\_9 		& 9	& 20	& 50		& 691		& 2.86e+19	& 11.6		& ---		& ---	\\
ran\_20\_60\_10		& 10	& 20	& 60		& 949		& 2.51e+22	& 20.3		& ---		& ---	\\
\hline
\end{tabular}
\end{table}

Table~\ref{table:comp} shows the results of comparison between volume estimation and computation.
``Dims'' represents the number of numeric variables in LACs.
``Ineqs'' represents the number of LACs and also represents the number of boolean variables in the boolean skeleton.
``Clauses'' represents the number of clauses in the boolean skeleton.
``Bunches'' represents the number of feasible partial assignments obtained by \texttt{VolCE}
and also represents the times of \texttt{VolCE} calling \texttt{PolyVest} or \texttt{Vinci}.

Observe that \texttt{VolCE} with \texttt{PolyVest} is very efficient and the relative errors of approximation are small.
When the dimension of instance grows to 8 or larger, \texttt{VolCE} with \texttt{Vinci} often fails to give an answer
in one hour or depletes memory.
Though ``Vinci'' has an option to restrict its memory storage, as a tradeoff it will take much more time to solve,
and still cannot solve instances within the timeout.

Table~\ref{table:comp0} shows the results of our tools with two-round strategy on random instances.
Column ``Sample'' represents the average coefficient of the sample size of $S_i$.
In the original algorithm, this average value always equals to $S = S_{max} = 1600$.
Values in column ``Ratio'' are approximation of saved sample points by the two-round strategy.
Obviously, the two-round strategy could save much time without losing much accuracy.
The differences of the results between original and two-round strategy are usually less than 5\%.
Besides, the two-round strategy could save 93\% to 97\% sample points and save more than 90\% time.

\begin{table}[!htbp]
\footnotesize \centering
\setlength{\abovecaptionskip}{0pt}
\setlength{\belowcaptionskip}{5pt}
\caption{Benefits of the Two-Round Strategy}\label{table:comp0}
\begin{tabular}{|c|c|c|c|c|c|c|c|c|c|}
\hline
\multicolumn{4}{|c|}{} & \multicolumn{2}{|c|}{Original}	& \multicolumn{3}{|c|}{Two-Round}	& \\
\hline
Dim.	& Ineq.	& Clause	& Bunch		& Result		& Time(s)	& Result	& Time(s)	& Sample	& Ratio\\
\hline
8		& 15	& 45		& 214		& 3.53e+17		& 31		& 3.58e+17	& 2.9		& 101.7		& 93.6\%	\\
8		& 20	& 60		& 480		& 1.10e+17		& 78.9		& 1.07e+17	& 6.3		& 75.5		& 95.3\%	\\
8		& 30	& 90		& 1135		& 6.94e+16		& 210.4		& 6.73e+16	& 20.7		& 54.7		& 96.6\%	\\
10		& 15	& 45		& 228		& 1.20e+23		& 68.5		& 1.20e+23	& 4.7		& 63.7		& 96.0\%	\\
10		& 20	& 60		& 949		& 2.52e+22		& 312.8		& 2.51e+22	& 20.3		& 61.5		& 96.2\%	\\
10		& 30	& 90		& 1394		& 8.06e+18		& 524.6 	& 7.92e+18	& 39.1		& 66.0		& 95.9\%	\\
15		& 40	& 200		& 1710		& 7.93e+27		& 2958.5	& 7.94e+27	& 189.1		& 44.6		& 97.2\%	\\
15		& 50	& 250		& 495		& 1.22e+23		& 984.4		& 1.23e+23	& 67.1		& 48.3		& 97.0\%	\\
20		& 40	& 200		& 8095		& ---			& ---		& 2.08e+40	& 2283		& 41.3		& 97.4\%	\\
20		& 60	& 400		& 689		& ---			& ---		& 6.71e+32	& 285		& 48.8		& 97.0\%	\\
30		& 60	& 400		& 886		& ---			& ---		& 6.59e+54	& 1528		& 44.0		& 97.3\%	\\
40		& 80	& 550		& 451		& ---			& ---		& 6.12e+66	& 2806		& 43.5		& 97.3\%	\\
\hline
\end{tabular}
\end{table}

\begin{table}[!htbp]
\footnotesize \centering
\setlength{\abovecaptionskip}{0pt}
\setlength{\belowcaptionskip}{5pt}
\caption{Experiments on Larger Instances}\label{table:larg}
\begin{tabular}{|c|c|c|c|c|c|c|c|}
\hline
Dims	& Ineqs	& Clauses	& Bunches	& Result		& Time(s)	\\
\hline
30	& 40	& 200		& 5200		& 1.42e+64		& 4314		\\
30	& 60	& 400		& 886		& 6.59e+54		& 1528		\\
30	& 80	& 500		& 610		& 2.01e+41		& 1227		\\
40	& 60	& 400		& 1752		& 5.23e+80		& 6437		\\
40	& 80	& 550		& 451		& 6.12e+66		& 2806		\\
40	& 100	& 750		& 111		& 5.00e+63		& 916		\\
50	& 100	& 750		& 24		& 5.63e+74		& 650		\\
\hline
\end{tabular}
\end{table}

Table~\ref{table:larg} shows the results of experiments with some larger random instances.
Note that the number of dimensions and bunches are the key parameters of the scale of instances.
The larger the number of dimensions or bunches, the more time the tool has to run.
\texttt{VolCE} can handle instances around 30-dimensions in reasonable time and up to 50-dimensions with a few bunches.

Table~\ref{table:explatte} are the results of counting integer solutions with \texttt{LattE}.
It shows that \texttt{LattE} can handle some problems up to 17 dimensions.
However, \texttt{LattE} cannot solve the random instance ``ran\_15\_45\_7''.
The inequalities in the instance are quite complicated.

\begin{table}[!htbp]
\footnotesize \centering
\setlength{\abovecaptionskip}{0pt}
\setlength{\belowcaptionskip}{5pt}
\caption{Experiments about counting integer solutions}\label{table:explatte}
\begin{tabular}{|c|c|c|c|c|c|}
\hline
Instance 	& Dims	& Ineqs	& Bunches	& Result	& Time(s)\\
\hline
abs\_0		& 1		& 1		& 1			& 128		& 0.000 \\
findmiddle\_2		
			& 3		& 6			& 2		& 5527040	& 0.004	\\
tritype\_16	& 4		& 18	& 4			& 8323072	& 0.051 \\
calDate\_13	
			& 6		& 5		& 1			& 7.99e+11	& 0.028	\\
Space\_manage\_1	
			& 17	& 3		& 1			& 2.69e+39	& 170.3	\\
bignum\_lia1	& 6	& 13	& 0			& 0		& 0.028	\\
bignum\_lia2	& 6	& 13	& 1			& 0		& 0.036	\\
SIMPLEBITADDER\_2
			& 12	& 51	& 98		& 0			& 1.3	\\
FISCHER1-1-fair	
			& 4		& 16	& 1			& 256		& 0.020	\\
FISCHER1-2-fair	
			& 6 	& 28	& 0			& 0			& 0.004	\\
prime\_cone\_sat\_2
			& 2		& 5		& 1			& 4159		& 0.004	\\
ran\_15\_45\_7		
			& 7		& 15	& 45		& ---		& ---	\\
\hline
\end{tabular} 
\end{table}

\section{Related Works}\label{sect:reltwks}

There was little work on the counting of SMT solutions, until quite recently.

Fredrikson and Jha \cite{FJ2014} relate a set of privacy and confidentiality
verification problems to the so-called {\it model-counting satisfiability\/} problem,
and present an abstract decision procedure for it.
They implemented this procedure for linear-integer arithmetic.
Their tool is called \verb!countersat!. It is not available to us.

Zhou {\it et al.} \cite{Zhou2015} propose a BDD-based search algorithm which reduces
the number of conjunctions.
For each conjunction, they propose a Monte-Carlo integration with a
ray-based sampling strategy, which approximates the volume.
Their tool is named \verb!RVC!. It can handle formulas with up to 18 variables.
But the running time is dozens of minutes.

A different approach is described in \cite{ChDM14}. It reduces an
approximate version of \verb!#SMT! to \verb!SMT!.
The approach does not need to modify existing SMT solvers.
It has been applied to solve a value estimation problem for
certain kind of probabilistic programs.
We do not know how large the benchmarks are,
and it is not clear about the quality of the approximation.

\section{Concluding Remarks}\label{sect:conclude}

\texttt{VolCE} is a tool for computing and estimating the volume of the solution space (or counting
the number of solutions), given a formula/constraint which is a Boolean combination
of linear arithmetic inequalities. \texttt{VolCE} is very flexible to use.
For medium sized SMT(LA) formulas, it can provide exact volume computation results
or exact number of solutions.
For larger SMT(LA) formulas, it can quickly perform volume estimation with high accuracy,
due to the use of effective heuristics.
We believe that the tool will be useful in a number of domains, such as program analysis and
probabilistic verification.

\appendix

\section{Introduction of \texttt{VolCE}}
\subsection{What is \texttt{VolCE}?}
\texttt{VolCE} is designed for computing or estimating the size of the solution space of an SMT formula
where the theory $T$ is restricted to the linear arithmetic theory. (SMT stands for Satisfiability
Modulo Theories.)
The prototype tool presented in~\cite{MLZ09} computes the exact volume of the solution space.
However, exact volume computation in general is an extremely difficult problem.
It has been proved to be \#P-hard, even for explicitly described polytopes.
On the other hand, it suffices to have an approximate value of the volume in many cases.
Later we implemented a tool to estimate the volume of polytopes~\cite{polyvest}
and integrated it into the framework of~\cite{MLZ09}.
The new tool is called \texttt{VolCE}.
It can efficiently handle instances of dozens of dimensions with high accuracy.
In addition, \texttt{VolCE} also accepts constraints involving independent Boolean variables.

\subsection{What can \texttt{VolCE} do?}
\texttt{VolCE} uses the following three packages:
\begin{itemize}
\item \texttt{PolyVest}~\cite{polyvest},
which can be used to estimate the volume of polytopes
\item \texttt{Vinci}~\cite{bib_vinci}, a software package that
implements several algorithms for (exact) volume computation.
\item
\texttt{LattE} (Lattice point Enumeration)~\cite{bib_latte}, a software package
dedicated to the problems of
counting lattice points and integration inside convex polytopes.
\end{itemize}

\section{Installing \texttt{VolCE}}



\begin{itemize}
\item{Step 1}:
Make sure that g++ (version 4.8 or higher version) is installed on your machine
(you can type ``\texttt{g++ -v}'' to check this).

\item{Step 2}:
The functionality of \texttt{VolCE} is dependent on some other libraries:
\texttt{zlib}, \texttt{boost}, \texttt{lpsolve}, \texttt{glpk}, \texttt{gfortran}, \texttt{LAPACK}, \texttt{BLAS} and \texttt{Armadillo}.
On Ubuntu or Debian, you can use ``apt-cache search'' and ``apt-get install'' to find and install all of these libraries.
Or you can download these libraries from:
\begin{table}[!htbp]
\centering
\begin{tabular}{|c|p{0.6\textwidth}|}
\hline
Library			& URL  \\
\hline
\texttt{zlib}	& http://zlib.net/\\
\hline
\texttt{boost}	& http://www.boost.org/\\
\hline
\texttt{lpsolve}	& http://sourceforge.net/projects/lpsolve/\\
\hline
\texttt{glpk}	& http://www.gnu.org/software/glpk/\\
\hline
\texttt{gfortran}	& http://gcc.gnu.org/fortran/\\
\hline
\texttt{LAPACK}	& http://www.netlib.org/lapack/\\
\hline
\texttt{BLAS}	& http://www.netlib.org/blas/index.html\\
\hline
\texttt{Armadillo} & http://arma.sourceforge.net/\\
\hline
\end{tabular}
\end{table}

\paragraph{Note.}
For \texttt{lpsolve},
make sure that its header files and the dynamic library (i.e., \texttt{liblpsolve55.so})
are included in the directories ``\texttt{/usr/include/lpsolve}'' and
``\texttt{/usr/lib/lp\_solve}'', respectively.
Besides, \texttt{LAPACK} and \texttt{BLAS} should be installed before installing \texttt{Armadillo}.

\item{Step 3}:
Open a shell (command line), change into the directory that was created by unpacking the \texttt{VolCE} archive, and type:
\begin{verbatim}
sh build.sh
\end{verbatim}

When the build process is finished, you will find all binaries in the directory ``\texttt{release/}''.

\item{Step 4}:
Install \texttt{LattE}~\cite{bib_latte}.
Build and move the executable files (\texttt{count} and \texttt{scdd\_gmp}) into directory ``\texttt{release/bin/}''.

\end{itemize}

This release of \texttt{VolCE} has been successfully built on the following operating systems:
\begin{itemize}
\item Ubuntu 12.04 on 64-bit with g++ 4.8.1
\item Ubuntu 13.10 on 32-bit with g++ 4.8.1
\end{itemize}

\section{Input Format}
The input of \texttt{VolCE} is an SMT formula where the theory $T$ is
restricted to the linear arithmetic theory.
It involves variables of various types (including integers, reals and Booleans).
We usually use $b_i$ to denote Boolean variables, $x_j$ to denote numeric variables.
In the input formula, there can be logical operators (like AND, OR, NOT),
arithmetic operators (like addition, subtraction, scalar multiplication)
and comparison operators (like $<$, $\le$, $>$, $\ge$, $=$, $\neq$).

Syntactically, there are two formats for the input file:
\begin{itemize}
\item \texttt{VolCE} style
\item \texttt{SMT-LIBv2}~\cite{SMTLIB}
\end{itemize}
We now describe them in detail.

\subsection{\texttt{VolCE} Style Input Format}

Let us introduce some concepts first.
\begin{itemize}
\item $LAC$: A linear arithmetic constraint (LAC) is a comparison between two linear arithmetic expressions.
Such a constraint can be denoted by a Boolean variable (e.g., $b_3 \equiv x_1 + x_2 \le 1$).
\item $literal$: A literal is either a Boolean variable (e.g., $b_3$) or
 a negated Boolean variable (e.g., NOT $b_3$).
\item $clause$: A clause is a set of one or more literals, connected with OR.
 (Boolean variables may not be repeated inside a clause.)
\item $formula$: A formula is a set of one or more clauses, connected with AND.
\end{itemize}

It is well known that any Boolean expression can be converted into the
conjunctive normal form (CNF) easily
(e.g., using the Tseitin transformation~\cite{TSEITIN}).
So the input of \texttt{VolCE} is a formula in the CNF form,
where each Boolean variable may stand for some LAC.
The input file generally consists of two parts: LACs, and clauses in the CNF.

An example of \texttt{VolCE} style formula is the following:
\begin{equation}\label{f:cnf1}
\begin{split}
& b_1 \equiv x_1 < x_2,\\
& b_3 \equiv x_1 + x_2 < 1,\\
& b_4 \equiv x_1 \le 1,\\
& b_5 \equiv x_2 \le 1,\\
& b_6 \equiv x_1 \ge 0,\\
& b_7 \equiv x_2 \ge 0,\\
& (b_1\ \mathrm{OR}\ (\mathrm{NOT}\ b3))\ \mathrm{AND}\\
& (b_1\ \mathrm{OR}\ (\mathrm{NOT}\ b2)\ \mathrm{OR}\ b3)\ \mathrm{AND}\\
& ((\mathrm{NOT}\ b3)\ \mathrm{OR}\ b4)\ \mathrm{AND}\\
& b_4\ \mathrm{AND}\ b_5\ \mathrm{AND}\ b_6\ \mathrm{AND}\ b_7.\\
\end{split}
\end{equation}

There are 7 Boolean variables ($b_1$, $\dots$, $b_7$),
2 numeric variables ($x_1$ and $x_2$), 6 LACs and 7 clauses in Formula~\ref{f:cnf1}.
Note that $b_2$ is an independent Boolean variable which does not represent any LAC.

Syntactically, \texttt{VolCE} accepts input in an ``Enhanced DIMACS CNF Format''.
Every line beginning with ``\verb!c!'' is a comment.
The first non-comment line must be of the form:
\begin{verbatim}
p cnf v lc BOOLS CLAUSES NUMVARS LACS
\end{verbatim}
It specifies the number of Boolean variables, the number of clauses, the number of
numeric variables and the number of linear constraints.

Every line beginning with ``\verb!m!'' defines a linear constraint and its corresponding Boolean variable.
It must be of the form:
\begin{verbatim}
m i a1 ... an op b
\end{verbatim}
It defines a linear inequality $a_1x_1 + \dots + a_nx_n\ op\ b$,
where $a_1$, $\dots$, $a_n$, $b$ are constants, and
$op$ is a comparison operator: $<$, $<=$, $>$, $>=$ or $=$.
(The tool does not support $\neq$ directly. However, $ax\neq b\ \Leftrightarrow\ \mathrm{NOT}\ ax=b$.)
The number $i$ means the Boolean variable $b_i$ represents this inequality.
The space between the character ``\verb!m!'' and the number $i$ is not mandatory.

Each of the other lines defines a clause:
a positive literal is denoted by the corresponding number (so \verb!4! means $b_4$),
and a negative literal is denoted by the corresponding negative number (so \verb!-5! means NOT $b_5$).
The last number in the line should be zero.
Each of these lines is a space-separated list of numbers.

So the above Formula~\ref{f:cnf1} would be written in the following way:
\begin{verbatim}
c It is an example, f1.vs.
p cnf v lc 7 7 2 6
c Linear Constraints part.
m1 1 -1 < 0
m3 1 1 < 1
m4 1 0 <= 1
m5 0 1 <= 1
m6 1 0 >= 0
m7 0 1 >= 0
c CNF part.
1 -3 0
1 -2 3 0
-1 3 0
4 0
5 0
6 0
7 0
\end{verbatim}
See the file \texttt{examples/f1.vs}.

\subsection{The \texttt{SMT-LIBv2} Language Inputs}
\texttt{VolCE} also partially supports the \texttt{SMT-LIBv2} language.
For details of this language, visit the website:
\begin{verbatim}
http://www.smt-lib.org/
\end{verbatim}

\texttt{VolCE} recognizes \texttt{SMT-LIBv2} format from the file name extension ``\texttt{.smt2}''.
It automatically parses such a file into the \texttt{VolCE} style input.

Table~\ref{table:supsmt2} lists the commands, variable types and identifiers of \texttt{SMT-LIBv2} language
that supported by \texttt{VolCE}.
\texttt{VolCE} ignores some basic commands like \texttt{set-logic}, \texttt{set-info}, \texttt{check-sat}, \texttt{exit}.
It directly checks all of the assertions.
Besides, \texttt{assert} commands must be written after all the \texttt{declare-fun} commands.

\begin{table}[!htbp]
\newcommand{\tabincell}[2]{\begin{tabular}{@{}#1@{}}#2\end{tabular}}
\centering
\setlength{\abovecaptionskip}{0pt}
\setlength{\belowcaptionskip}{5pt}
\caption{Supported SMT-LIBv2 Components}\label{table:supsmt2}
\begin{tabular}{|c|p{0.5\textwidth}|}
\hline
Commands  & \texttt{declare-fun} \texttt{assert} \\
\hline
Variable Types  & \texttt{Int} \texttt{Real} \texttt{Bool} \\
\hline
Identifiers  & \tabincell{c}{\texttt{let} \\
		\texttt{and} \texttt{or} \texttt{not} $\texttt{=>}$ \texttt{ite} \\
		$\texttt{+}$ $\texttt{-}$ $\texttt{*}$ $\texttt{/}$ \\
		$\texttt{=}$ $\texttt{>}$ $\texttt{>=}$ $\texttt{<}$ $\texttt{<=}$ \texttt{distinct}} \\
\hline
\end{tabular}
\end{table}

In the \texttt{SMT-LIBv2} language, the above Formula~\ref{f:cnf1} would be written like this:
\begin{verbatim}
(set-logic QF_LRA)
(set-info :f1.smt2)
(set-info :smt-lib-version 2.0)
(set-info :status sat)
(declare-fun x () Real)
(declare-fun y () Real)
(declare-fun b () Bool)
(assert (and (<= x 1) (<= y 1) (>= x 0) (>= y 0)))
(assert (let ((v1 (< (+ x y) 1)) (v2 (< x y)))
(and (or v1 (not v2)) (or v1 v2 b) (or (not v1) v2))))
(check-sat)
(exit)
\end{verbatim}
See the file \texttt{examples/f1.smt2}.

\section{Running \texttt{VolCE}}

To run \texttt{VolCE}, you should switch your working directory to the absolute path of \texttt{VolCE}.

\texttt{VolCE} has a help menu. To view it, simply type the command \verb!"./volce --help"!.

The general usage of \texttt{VolCE} is
\begin{verbatim}
  % ./volce [OPTION]... <INPUT-FILE>
\end{verbatim}

The meanings of the options are given in the following table.

\begin{longtable}{|c|p{0.8\textwidth}|}
\caption{Command-line Options of \texttt{VolCE}}\label{table:cmdopt}\\
\hline
Option & Meaning  \\
\hline
\texttt{-P}  &
Enables \texttt{PolyVest} for volume approximation. The input variables in the linear inequalities are reals.
By default, \texttt{VolCE} calls \texttt{PolyVest}. It assumes that all the numeric variables are reals.
 \\
\hline
\texttt{-V}  &
Enables \texttt{Vinci} for volume computation. The input variables in the linear inequalities are reals.
 \\
\hline
\texttt{-L}  &
Enables \texttt{LattE} to count the number of integer solutions.
The input variables in the linear inequalities should be integers.
This option is usually enabled in the case of integer variables.
 \\
\hline
\texttt{-w=NUMBER} &
Specifies the word length of numeric variables in bit-wise representations.
Then each variable is automatically bounded by the range $[-2^{w-1}, 2^{w-1}-1]$.
Setting the word length to 0 will disable this feature.
By default, the word length is 8, which means the domain of every numeric variable is $[-128, 127]$.
For example, you can change it to 3, by using the option \texttt{-w=3}.
 \\
\hline
\texttt{-maxc=NUMBER} &
Sets the maximum sampling coefficient of \texttt{PolyVest}, which is an upper bound.
Generally, the larger this coefficient is, the more accurate the result will be.
However, the running time of the tool will be longer.
The default value of \texttt{maxc} is 1600.
For example, you may change it to 1000, by using the option \texttt{-maxc=1000}.\\
\hline
\texttt{-minc=NUMBER} &
Sets the minimum sampling coefficient of \texttt{PolyVest}.
The default value is 40. \\
\hline
\end{longtable}

To estimate the volume of the solution space of Formula~\ref{f:cnf1}, simply type:
\begin{verbatim}
  % ./volce examples/f1.vs
\end{verbatim}

Note that Formula~\ref{f:cnf1} guarantees $0\le x_1, x_2\le 1$.
So we can disable the internal bit-wise bounds of numeric variables, by setting the word length to 0:
\begin{verbatim}
  % ./volce -w=0 examples/f1.vs
\end{verbatim}

You can also enable \texttt{PolyVest}, \texttt{Vinci}, \texttt{LattE} at the same time:
\begin{verbatim}
  % ./volce -P -V -L -w=0 examples/f1.vs
\end{verbatim}

\paragraph{Remarks}
Several tools (\texttt{PolyVest}, \texttt{Vinci}, \texttt{LattE}) have been integrated
which can be enabled for different situations.
\texttt{Vinci} gives an accurate volume for a polytope; but it may
have difficulty handling problem instances with more than 10 numeric variables.
\texttt{PolyVest} gives approximate results, but it can deal with larger instances.
\texttt{LattE} is good at counting the number of integer solutions.
Sometimes, the first two tools can also be used for approximating the number of integer solutions.

\section{Examples}
\paragraph{Example 1}
For the above Formula~\ref{f:cnf1}, we have two input files:
\texttt{VolCE} style input (\texttt{f1.vs}) and \texttt{SMT-LIBv2} input (\texttt{f1.smt2}).

Execute the command:
\begin{verbatim}
  % ./volce -P -V -L -w=0 examples/f1.smt2
\end{verbatim}

And we obtain the result:
\begin{multicols}{2}
\begin{verbatim}
Enabled PolyVest.
Enabled Vinci.
Enabled LattE.
Set word length to 0.
Disabled default bounds since wo
rd length <= 0.
VolCE Directory: ...
Working Directory: ...

================================

Parsing smt2 file.
Reading Input.
Number of bool vars:     16
Number of clauses:       29
Number of numeric vars:  2
Number of linear constraints:  6

================================

Branches: 2
SATISFIABLE

================================
=========== PolyVest ===========
================================

FIRST ROUND
0	0.222875 * 2
1	0.24338 * 1

SEC & LAST ROUND
0	1600	0.252037 * 2
1	1600	0.250964 * 1

Total approximation: 0.755039

================================

The total volume (PolyVest): 0.7
5503900


================================
============ Vinci =============
================================

0.25000000 * 2
0.25000000 * 1

================================

The total volume (Vinci): 0.7500
0000


================================
============ LattE =============
================================

0 * 2
2 * 1

================================

The total volume (LattE): 2

\end{verbatim}
\end{multicols}

Analysis:

\begin{figure}[!htbp]
\centering
\includegraphics[width=0.5\textwidth]{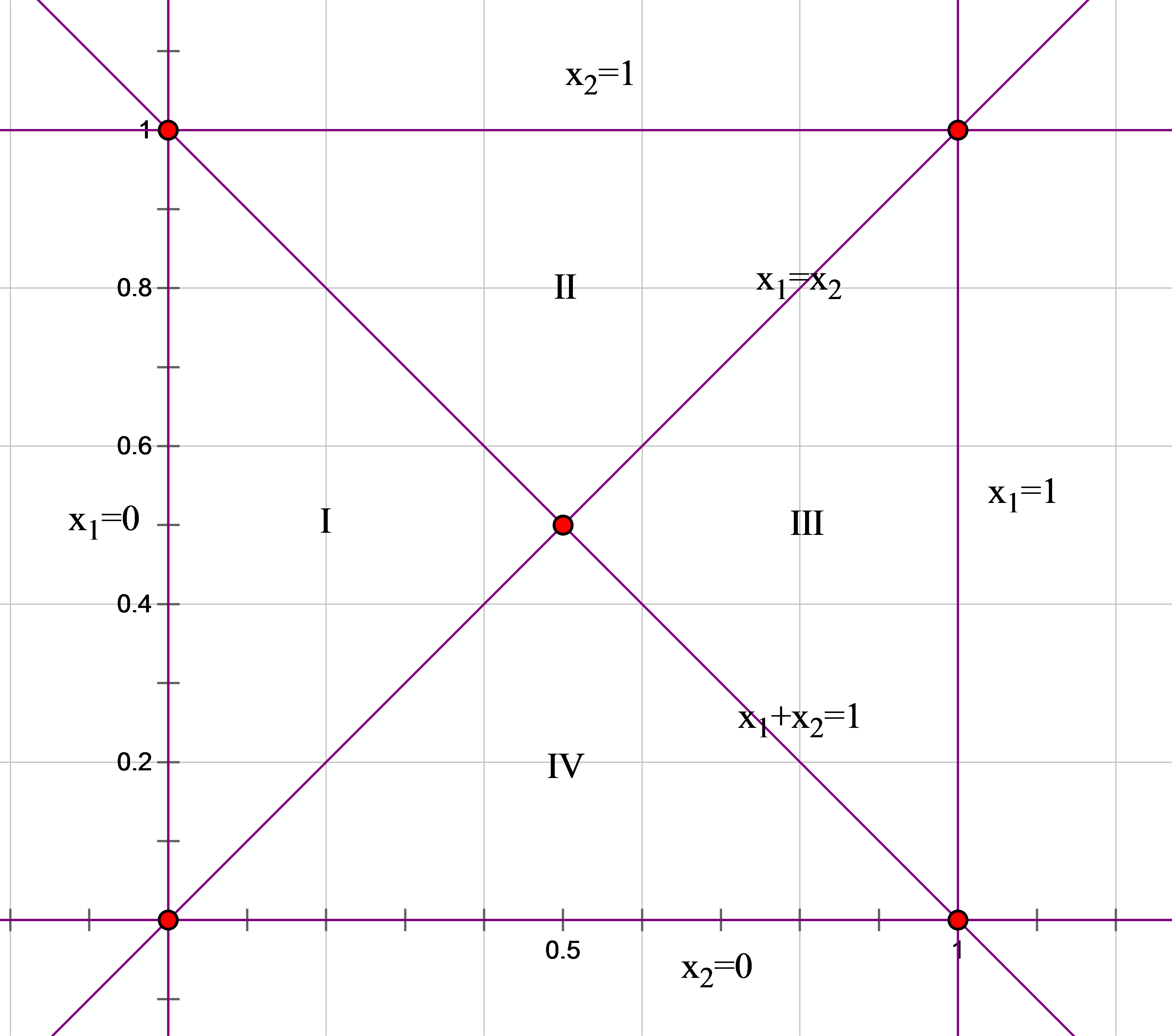}
\caption{Solution Space of Formula~\ref{f:cnf1}}\label{fig:1}
\end{figure}

Figure~\ref{fig:1} shows the linear constraints in Formula~\ref{f:cnf1}.
The plane is splitted into 4 areas, since $b_4$, $b_5$, $b_6$, $b_7$ are always True.
The pair \{$b_1$, $b_3$\} determines the counted areas.
\begin{itemize}
\item {Area I}: \{$b_1$ = True, $b_3$ = True\}. It has no lattice points.
\item {Area II}: \{$b_1$ = True, $b_3$ = False\}. It has 1 lattice point: \{0, 1\}.
\item {Area III}: \{$b_1$ = False, $b_3$ = False\}. It has 2 lattice points: \{1, 0\} and \{1, 1\}.
\item {Area IV}: \{$b_1$ = False, $b_3$ = True\}. It has 1 lattice point: \{0, 0\}.
\end{itemize}
There are 3 Boolean solutions for Formula~\ref{f:cnf1}:
\{$b_1$ = True, $b_2$ = True, $b_3$ = True\},
\{$b_1$ = True, $b_2$ = False, $b_3$ = True\}, and
\{$b_1$ = False, $b_2$ = True, $b_3$ = False\}.
Thus the volume of the solution space is $2 \times vol(Area_I) + vol(Area_{III}) = 0.75$.
And there are $2 \times 0 + 2 = 2$ integer solutions (lattice points).

\paragraph{Example 2}
Here is an exercise for young pupils: In the following square, there are 8 sub-areas.
Color them so that the neighboring sub-areas use different colors.
How many different coloring schemes are there?
\begin{center}
\includegraphics[height=4cm]{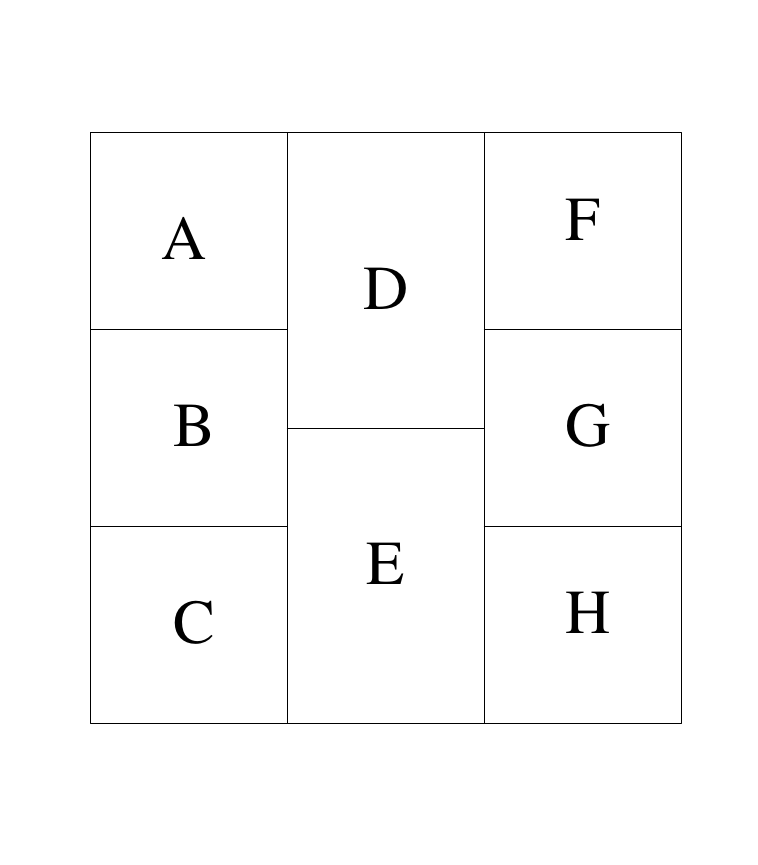}
\end{center}
Obviously, this problem can be regarded as a solution counting problem.
The input consists of the following inequalities:
\[ xA \neq xB, ~~ xA \neq xD, ~~ xB \neq xC, ~~ xB \neq xD, ~~ xB \neq xE, \]
\[ xC \neq xE, ~~ xD \neq xE, ~~ xD \neq xF, ~~ xD \neq xG, \]
\[ xE \neq xG, ~~ xE \neq xH, ~~ xF \neq xG, ~~ xG \neq xH. \]
We assume that there are at most 4 colors, and execute the following command:
\begin{verbatim}
  % ./volce -L -w=2 examples/coloring.smt2
\end{verbatim}
We find that there are 768 solutions.

\paragraph{Example 3}
In~\cite{MLZ09}, we describe a program called {\it getop()}
and analyze the execution frequency of its paths.
For {\it Path1}, its path condition\footnote{The path condition
is a set of constraints such that any input data satisfying
these constraints will make the program execute along that path.}
is:
\begin{verbatim}
  (NOT ((c = 32) OR (c = 9) OR (c = 10))) AND
  ((c != 46) AND ((c < 48) OR (c > 57)))
\end{verbatim}
Here {\tt c} is a variable of type {\tt char}; it can be
regarded as an integer variable within the domain [-128..127].

For the above path condition, we can compute the number of solutions
by executing the command:
\begin{verbatim}
  % ./volce -L examples/program_analysis/getopPath1.smt2
\end{verbatim}
We find that the path condition has $242$ solutions.
(We do not need to use the option {\tt -w=8},
because the default word length is $8$.)

Given that the size of the whole search space is $256$, we conclude
that the frequency of executing {\it Path1} is about $0.945$ (i.e., $242/256$).
This means, if the input string has only one character, most probably,
the program will follow this path.

Another path, {\it Path2}, has the following path condition:
\begin{verbatim}
  ((c0 = 32) OR (c0 = 9) OR (c0 = 10)) AND
  (NOT ((c1 = 32) OR (c1 = 9) OR (c1 = 10))) AND
  (NOT ((c1 != 46) AND ((c1 < 48) OR (c1 > 57)))) AND
  (NOT ((c2 >= 48) AND (c2 <= 57))) AND
  (NOT (c2 = 46))
\end{verbatim}
Given this set of constraints, and using {\tt LattE},
our tool tells us that the number of solutions is 8085.
The executed command is:
\begin{verbatim}
  % ./volce -L examples/program_analysis/getopPath2.smt2
\end{verbatim}
So the path execution frequency is  $8085 / (256*256*256)$
which is roughly 0.00048.

\paragraph{Example 4}
Hoare's program FIND takes an array A[N] and an integer
as input, and partitions the array into two parts.

Assume that $N = 8$.
We may extract two execution paths from the program, and generate the path conditions.
The first path condition is the following:
\begin{verbatim}
  (A[0] < A[3]);  !(A[1] < A[3]);  (A[3] < A[7]);
  !(A[3] < A[6]);  !(A[2] < A[3]);  !(A[3] < A[5]);
  !(A[3] < A[4]);  (A[0] < A[4]);  (A[6] < A[4]);  (A[5] < A[4]).
\end{verbatim}
Setting the word length to 4, we can find that the number of solutions is $4075920$.
The executed command is:
\begin{verbatim}
  % ./volce -L -w=4 examples/program_analysis/FINDpath1.smt2
\end{verbatim}

The second path condition is a bit more complicated:
\begin{verbatim}
  !(A[0] < A[3]);  (A[3] < A[7]);  (A[3] < A[6]);
  (A[3] < A[5]);  (A[3] < A[4]);  !(A[1] < A[3]);
  (A[3] < A[2]);  (A[3] < A[1]);  (A[1] < A[0]);
  (A[2] < A[0]);  !(A[0] < A[7]);  !(A[4] < A[0]);
  (A[0] < A[6]);  !(A[0] < A[5]);  (A[1] < A[7]);
  (A[2] < A[7]);  !(A[7] < A[5]);  !(A[1] < A[5]);
  !(A[2] < A[5]);  (A[5] < A[2]);  (A[2] < A[1]).
\end{verbatim}
Executing the command:
\begin{verbatim}
  % ./volce -L -w=4 examples/program_analysis/FINDpath2.smt2
\end{verbatim}
we find that the number of solutions is $87516$.
So, the first path is executed much more frequently than the second one.
(We assume that the input space is evenly distributed.)

\paragraph{Example 5}
Let us try a randomly generated example (\texttt{ran\_5\_20\_8.vs}).
It has 5 Boolean variables, 8 numeric variable, 20 clauses and 5 linear constraints.

Execute the command:
\begin{verbatim}
  % ./volce -P -V -w=4 examples/ran/ran_5_20_8.vs
\end{verbatim}

And we obtain the result:
\begin{multicols}{2}
\begin{verbatim}
Enabled PolyVest.
Enabled Vinci.
Set word length to 4.
VolCE Directory: ...
Working Directory: ...

================================

Reading Input.
Number of bool vars:     5
Number of clauses:       20
Number of numeric vars:  8
Number of linear constraints:  5

================================

Branches: 1
SATISFIABLE

================================
=========== PolyVest ===========
================================

FIRST ROUND
0	8.38972e+06 * 1

SEC & LAST ROUND
0	1600	7.88093e+06 * 1

Total approximation: 7.88093e+06

================================

The total volume (PolyVest): 788
0930.00000000


================================
============ Vinci =============
================================

7970738.22355500 * 1

================================

The total volume (Vinci): 797073
8.22355500
\end{verbatim}
\end{multicols}


\begin{thebibliography}{9}

\bibitem{Smith:1987}
H.C.P. Berbee, C.G.E. Boender, A.H.G. Rinnooy Ran, C.L. Scheffer, R.L. Smith and J. Telgen.
\newblock Hit-and-run algorithms for the identification of nonredundant linear inequalities.
\newblock \emph{Mathematical Programming}, 37(2): 184--207 (1987).

\bibitem{lpsolve}
M. Berkelaar {\em et al.} \newblock lp\_solve.
\verb!http://lpsolve.sourceforge.net/!

\bibitem{vinci98}
B. B{\"u}eler, A. Enge and K. Fukuda.
\newblock Exact volume computation for polytopes: a practical study.
\newblock In {\it Polytopes -- combinatorics and computation}, 1998.
\newblock Available at \\
\verb!http://www.math.u-bordeaux1.fr/~enge/index.php?category=software&page=vinci!

\bibitem{ChDM14}
D. Chistikov, R. Dimitrova and R. Majumdar.
\newblock Approximate counting in {SMT} and value estimation for probabilistic programs, Nov 2014.
\newblock \verb!http://arxiv.org/abs/1411.0659!

\bibitem{latte04}
J.A. De Loera {\em et al.}
\newblock Effective lattice point counting in rational convex polytopes.
\newblock {\it J. of Symbolic Computation}, 38(4): 1273--1302 (2004).

\bibitem{Kannan:1989}
M. Dyer, A. Frieze and R. Kannan.
\newblock A random polynomial time algorithm for approximating the volume of convex bodies.
\newblock \emph{Proc. ACM SToC}, pp.375--381 (1989)

\bibitem{FJ2014}
M. Fredrikson and S. Jha.
\newblock Satisfiability modulo counting: a new approach for analyzing privacy properties.
\newblock {\it Proc. CSL-LICS'14}, Article No. 42 (2014).

\bibitem{polyvest}
C. Ge, F. Ma and J. Zhang.
\newblock A fast and practical method to estimate volumes of convex polytopes, Dec 2013.
\newblock \verb!http://arxiv.org/abs/1401.0120/!

\bibitem{GDV12}
J. Geldenhuys, M.B. Dwyer and W. Visser.
\newblock Probabilistic symbolic execution.
\newblock {\it Proc. ISSTA 2012}, pp.166--176.

\bibitem{Book:1993}
M. Gr{\"o}tschel, L. Lov{\'a}sz and A. Schrijver.
\newblock \emph{Geometric Algorithms and Combinatorial Optimization}.
\newblock Springer Verlag (1993)

\bibitem{LZ11}
S. Liu and J. Zhang.
\newblock Program analysis: from qualitative analysis to quantitative analysis.
\newblock {\it Proc. ICSE 2011 (NIER track)}, pp.956--959.

\bibitem{Lovasz:2006}
L. Lov{\'a}sz and S. Vempala.
\newblock Simulated annealing in convex bodies and an $O^*(n^4)$ volume algorithm.
\newblock \emph{J. of Computer and System Sci.}, 72(2): 392--417 (2006)

\bibitem{MLZ09}
F. Ma, S. Liu and J. Zhang.
\newblock Volume computation for Boolean combination of linear arithmetic constraints.
\newblock {\it Proc. CADE-22}, LNCS 5663, pp.453--468, 2009.

\bibitem{bib_latte}
LattE, available at ~
\verb!https://www.math.ucdavis.edu/~latte/!

\bibitem{SMTLIB}
SMT-LIB: The Satisfiability Modulo Theories Library. ~ \verb!http://www.smt-lib.org/!

\bibitem{TSEITIN}
G.S. Tseitin. On the complexity of derivation in propositional calculus.
In: {\it Slisenko, A.O. (ed.) Structures in Constructive Mathematics and
Mathematical Logic, Part II, Seminars in Mathematics (translated from Russian)},
pp.115--125, Steklov Mathematical Institute, 1968.

\bibitem{bib_vinci}
Vinci, available at \\
\verb!http://www.math.u-bordeaux1.fr/~aenge/?category=software&page=vinci!

\bibitem{Zhou2015}
M. Zhou, F. He, X. Song, S. He, G. Chen and M. Gu.
\newblock Estimating the volume of solution space for satisfiability modulo linear real arithmetic.
\newblock \emph{Theory of Computing Systems}, 56(2): 347--371 (2015).

\end{thebibliography}
\end{document}